\makeatletter
\@ifundefined{isChecklistMainFile}{
  \newif\ifreproStandalone
  \reproStandalonetrue
}{
  \newif\ifreproStandalone
  \reproStandalonefalse
}
\makeatother

\ifreproStandalone

\documentclass[letterpaper]{article} 
\usepackage[draft]{aaai2026}  
\usepackage{times}  
\usepackage{helvet}  
\usepackage{courier}  
\usepackage[hyphens]{url}  
\usepackage{graphicx} 
\usepackage{natbib}  
\usepackage{caption} 
\usepackage{algorithm}
\usepackage{algorithmic}
\usepackage{newfloat}
\usepackage{listings}

\usepackage{multicol}
\usepackage{multirow}
\usepackage{booktabs}
\usepackage{amsmath,amsfonts}
\usepackage{algorithmic}
\usepackage{algorithm}
\usepackage{graphicx}
\usepackage{enumitem}
\usepackage{xspace}
\usepackage{caption}
\usepackage{xcolor}

\newcommand{\ie}{\emph{i.e., }}
\newcommand{\eg}{\emph{e.g., }}
\newcommand{\etal}{\emph{et al.}}

\DeclareCaptionStyle{ruled}{labelfont=normalfont,labelsep=colon,strut=off} 
\floatstyle{ruled}
\newfloat{listing}{tb}{lst}{}
\floatname{listing}{Listing}
\title{Delayed Feedback Modeling with Influence Functions}

\author{
Chenlu Ding\textsuperscript{\rm 1}, 
Jiancan Wu\textsuperscript{\rm 1}, 
Yancheng Yuan\textsuperscript{\rm 2}, 
Cunchun Li\textsuperscript{\rm 1}, 
Xiang Wang\textsuperscript{\rm 1}, 
Dingxian Wang\textsuperscript{\rm 3}, 
Frank Yang\textsuperscript{\rm 3}, 
Andrew Rabinovich\textsuperscript{\rm 3}
}

\affiliations{
\textsuperscript{\rm 1}University of Science and Technology of China \quad
\textsuperscript{\rm 2}The Hong Kong Polytechnic University \quad
\textsuperscript{\rm 3}Upwork \\
\texttt{dingchenlu200103@gmail.com}
}

\begin{document}
\maketitle

\begin{abstract}
In online advertising under the cost-per-conversion (CPA) model, accurate conversion rate (CVR) prediction is crucial. A major challenge is delayed feedback, where conversions may occur long after user interactions, leading to incomplete recent data and biased model training. Existing solutions partially mitigate this issue but often rely on auxiliary models, making them computationally inefficient and less adaptive to user interest shifts. We propose IF-DFM, an \underline{I}nfluence \underline{F}unction-empowered for \underline{D}elayed \underline{F}eedback \underline{M}odeling which estimates the impact of newly arrived and delayed conversions on model parameters, enabling efficient updates without full retraining. By reformulating the inverse Hessian-vector product as an optimization problem, IF-DFM achieves a favorable trade-off between scalability and effectiveness. Experiments on benchmark datasets show that IF-DFM outperforms prior methods in both accuracy and adaptability.
\end{abstract}

\section{Introduction}
In online advertising, predicting conversion rate (CVR) is vital for revenue optimization under the cost-per-conversion (CPA) model \cite{ma2018entire,lu2017practical,lee2012estimating}, where advertisers pay only after users complete specific actions. Unlike immediate clicks, conversions often occur hours or even weeks after the ad click, leading to the \emph{delayed feedback issue} \cite{chapelle2014modeling,ktena2019addressing,yang2021capturing}. To address this, platforms typically define a fixed time window to wait for conversions and periodically update the model \cite{dai2023dually}. Clicks followed by conversions within the window are treated as positives, while others --- including those with longer delays --- are marked negative \cite{guo2022calibrated,ktena2019addressing,li2021follow}. This creates a trade-off between label accuracy and model timeliness.

\begin{figure}[t]
    \centering
\includegraphics[width=0.47\textwidth]{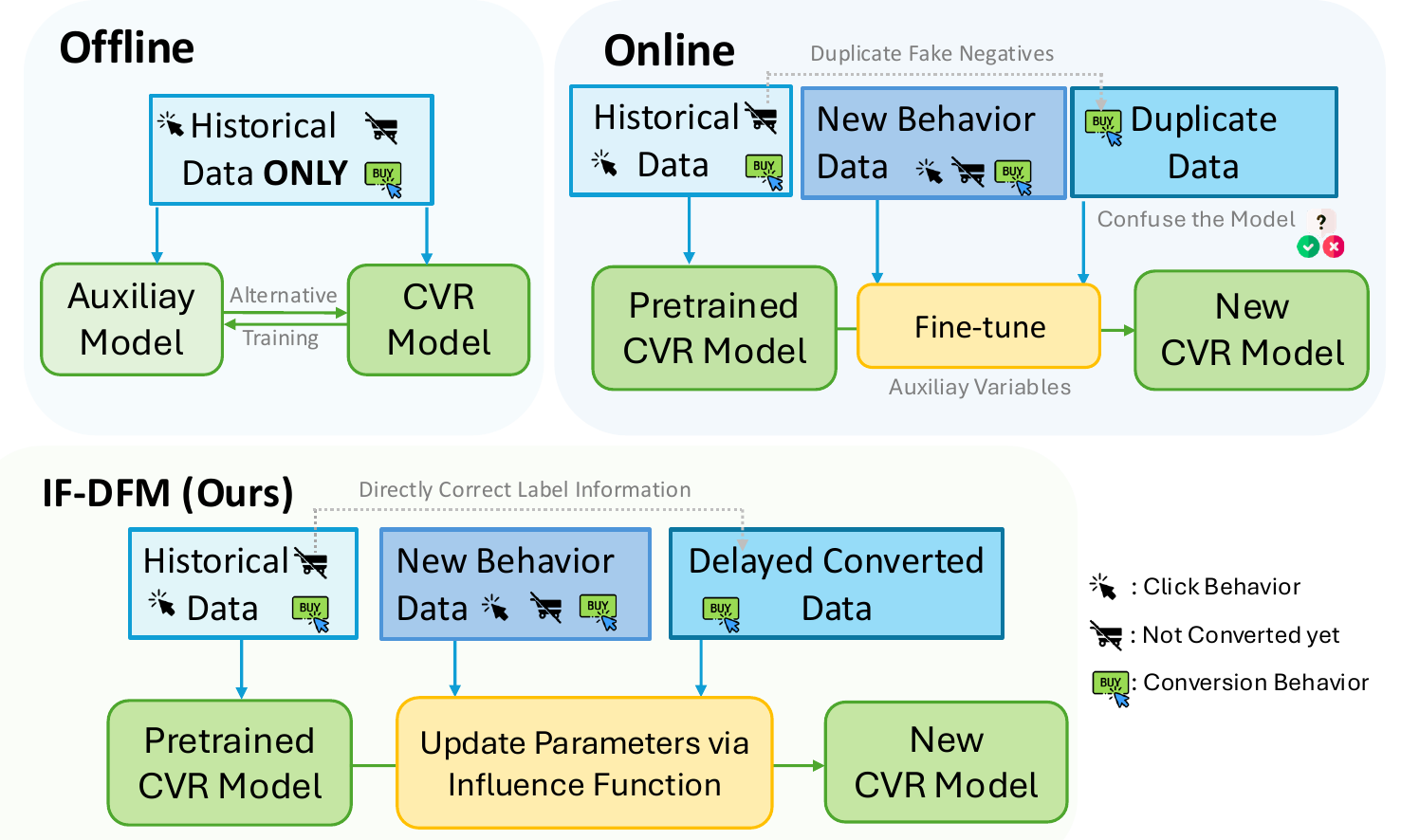}
    \caption{The framework of offline CVR methods, online CVR methods, and IF-DFM.}
    \label{fig:compare}
\end{figure}

Many efforts have been made to mitigate the delayed feedback issue in CVR, which can be roughly categorized into offline and online approaches.
Offline methods \cite{yoshikawa2018nonparametric,saito2020dual, yasui2020feedback} rely on historical data to train models, usually with an extra component to model the distribution of the delay time \cite{chapelle2014modeling} or adjust for the correct labels \cite{wang2023unbiased}. They generally work under the i.i.d assumption that future data distributions will remain consistent with historical patterns.
On the other hand, online methods \cite{chen2022asymptotically, gu2021real, li2021follow} attempt to update the model by ingesting newly observed data and correcting labels in near real-time or waiting short intervals. Such updates are conducted on the duplicated data with corrected labels \cite{yang2021capturing} or using a surrogate loss, to approach the oracle model \cite{dai2023dually,ktena2019addressing}. The framework of offline CVR methods, online
CVR methods can be found in Figure~\ref{fig:compare}.

Despite effectiveness, current approaches suffer from two significant limitations:
\begin{itemize}[leftmargin=*]
    \item \textbf{Inadequate Adaptation to Evolving User Interest.} {The dynamic nature of user interests challenges the assumption of an unchanged data distribution, which may not hold in environments characterized by evolving user behavior \cite{dai2023dually}.} Offline methods, operating under the i.i.d assumption, struggle to capture users' emerging behaviors and preference shifts. While online methods update with new data promptly, their reliance on duplicated samples for label correction can cause confusion and limit the effective use of accurate labels.
    \item \textbf{Reliance on Auxiliary Models.} The mainstream approaches typically incorporate auxiliary components to estimate potential label reversals for observed negatives or to model the probabilities of fake negatives as latent variables. However, developing these additional components can be as complex as constructing the primary CVR model. Moreover, their effectiveness is limited by the quality and quantity of historical data, resulting in computational inefficiencies and adding additional complexities to the CVR prediction task.
\end{itemize}

We explore a new paradigm for mitigating delayed feedback that naturally adapts to new data without auxiliary models. While retraining with correctly labeled data is straightforward, it is impractical in large-scale CVR settings due to high computational costs. To address this, we leverage influence functions---a tool from robust statistics gaining traction in machine learning~\cite{koh2017understanding}. Influence functions estimate parameter changes caused by data perturbations (\eg removing a sample~\cite{zhang2023recommendation} or editing features~\cite{wu2023gif}) by up-weighting the sample’s loss, resulting in inverse Hessian-vector products. 
Conceptually, incorporating the influence function to mitigate the delayed feedback problem offers a notable benefit: it allows for the direct modification of model parameters based on the estimated impact of mislabeled or newly injected data. However, given the significant computational and space complexities of inverting the Hessian matrix, integrating the influence function also presents its unique challenge: How to efficiently get model parameter changes for CVR tasks?
In this paper, we propose an \underline{i}nfluence \underline{f}unction-empowered framework for \underline{d}elayed \underline{f}eedback \underline{m}odeling (IF-DFM).
The core idea is to leverage influence functions to estimate the impact of newly injected data and directly update model parameters without retraining.
For a deployed CVR model, IF-DFM computes the parameter changes caused by correcting mislabeled samples, approximating the difference between the current model and the one trained on corrected data.
This avoids sample duplication and eliminates the need for auxiliary models.
Additionally, IF-DFM supports the integration of newly arrived behavior data (\eg, post-deployment clicks), enabling the model to adapt to evolving user interactions and improve prediction accuracy.
To reduce computation, we reformulate the inverse Hessian-vector product as a finite-sum quadratic problem, which can be efficiently solved using stochastic optimization methods like SGD~\cite{robbins1951stochastic} and its variants~\cite{duchi2011adaptive,kingma2014adam}, avoiding full-batch gradients.
We conduct extensive experiments on two benchmark datasets, demonstrating the superiority of the proposed method.
Our main contributions can be summarized as follows:

\begin{itemize}[leftmargin=*]
    \item We propose a new paradigm IF-DFM for mitigating the delayed feedback problem.
    It utilizes the influence function to estimate the impact of fake label correction and new behavior data integration, and directly modify model parameters without retraining.
    \item We propose to compute the influence function by formulating the inverse Hessian-vector-product calculation as an optimization problem, achieving a balance between computational efficiency and effectiveness.
    \item Extensive experiments on benchmark datasets demonstrate the superiority of IF-DFM.
\end{itemize}

\section{Preliminaries}
\subsection{Problem Formulation}
The fundamental task of CVR prediction is to estimate the rate of conversion from the user clicks into purchases, based on historical data. Throughout this paper, we represent the $i$-th training sample as $z_i = (x_i, y_i^{O})$, where $x_i$ encompasses the features, $y_i^{O}$ is the observed label. Furthermore, in CVR prediction, each sample is also associated with a true label $y_i^{T}$, which is inaccessible within the training dataset. The feature set $x_i$ typically comprises temporal components --- click timestamp $c_i$, payment timestamp $p_i$, and the elapsed time $e_i$ from the click to a model-training-start timestamp $T$  --- along with user/item characteristics.
Here, the default value of $p_i$ is $-1$ if no conversion has occurred before $T$. The training set accumulated up to $T$ can be described as $\mathcal{Z} = \{ z_i \,|\, i=1,2,...,n\}$, where $n$ is the sample size. Due to the delayed feedback, where the ground truth $y_i^{T}$ is absent, the training process could be misled by fake negative samples. 
\begin{figure}[t]
    \centering
    \includegraphics[width=0.48\textwidth]{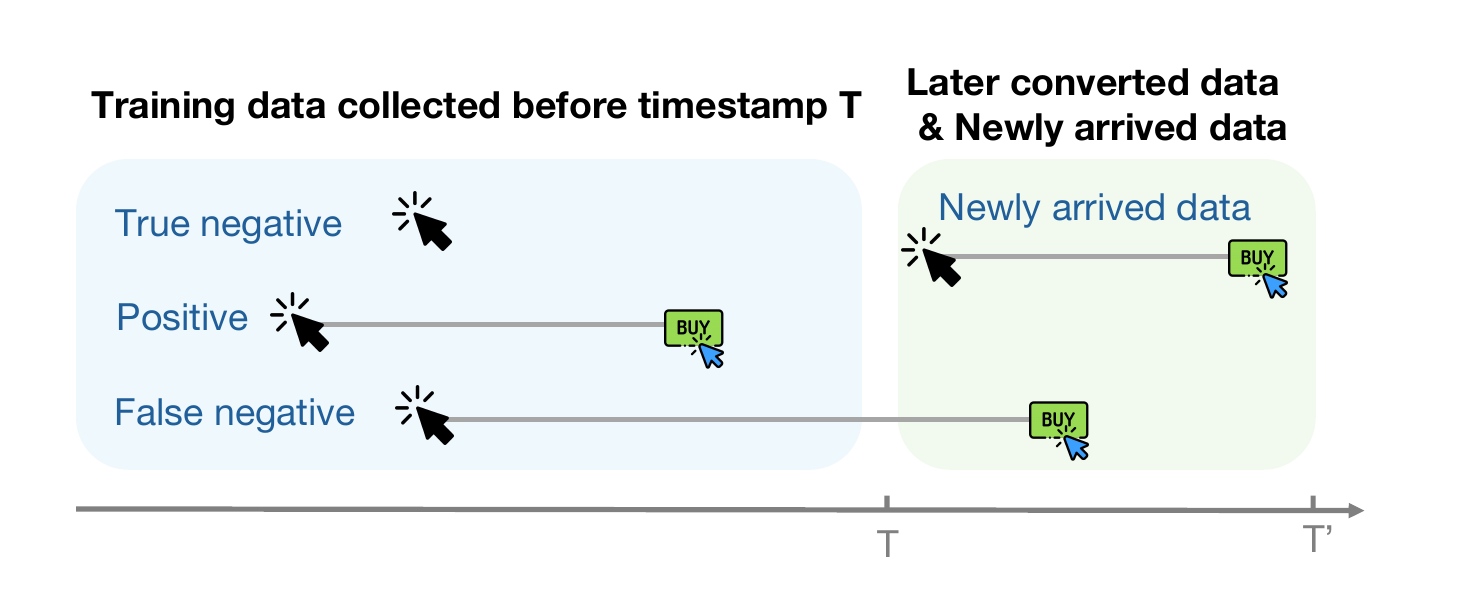}
    \caption{An illustration of the delayed feedback problem in CVR tasks.}
    \label{fig:demo}
\end{figure}

The relationship between sample labels and the timeline is illustrated in Figure~\ref{fig:demo}. If a conversion occurs before $T$ (\ie $p_i < T$), the sample is labeled as $y_i^{O} = y_i^{T} = 1$, representing a true positive. If the conversion is unobserved by $T$ (\ie $p_i > T$), it is labeled as $y_i^{O} = 0$, which may not match $y_i^T$, resulting in a fake negative due to delayed conversion. Thus, delayed feedback arises solely from negatives that may later become positives. The model is evaluated at timestamp $T'$, using only the test samples with their true labels.

\subsection{Vanilla and Retrain Versions of Loss}
The CVR model, $f(x,\theta)$ parameterized by $\theta$, operates as a binary classifier, predicting whether a click action $x$ will result in a conversion. Training such models typically employs binary cross-entropy loss \cite{dai2023dually,wang2023unbiased}. We begin with the \textbf{Vanilla} version~\cite{chapelle2014modeling,yasui2020feedback}, where all unconverted samples are treated as negatives:
\begin{equation}
\label{vanilla}
\mathcal{L}_{ {V }} (\theta) = \frac{1}{n} \sum_{i=1}^{n} \mathcal{L}_{BCE}(z_i, \theta),
\end{equation}
where for any sample $z = (x, y)$ and model parameter $\theta$,
$$\mathcal{L}_{BCE}(z, \theta) = - [ y \log f\left(x ,\theta\right)+\left(1-y\right) \log \left(1-f\left(x ,\theta\right)\right) ]$$ 
is the commonly used binary cross-entropy loss.

The vanilla loss is computed over observed labels without correction, serving as a lower bound for CVR performance. In contrast, the \textbf{Oracle} version assumes access to true labels $y_i^T$, providing an upper bound:
\begin{equation}
\label{oracle}
\mathcal{L}_{{O}} (\theta) = \frac{1}{n} \sum_{i=1}^{n} \mathcal{L}_{BCE}(z_i^T, \theta),
\end{equation}
where $z_i^T = (x_i, y_i^T)$. Since many conversions occur after model deployment, true labels are typically unavailable during training. To address this, we define a practical \textbf{Retrain} version:
\begin{equation}
\label{retrain}
\mathcal{L}_{{R}}(\theta) =\frac{1}{n} \sum_{i=1}^{n}\mathcal{L}_{BCE}(\tilde{z}_i, \theta),
\end{equation}
where $\tilde{z}_i = (x_i, \tilde{y_i})$ and $\tilde{y}_i$ is defined as:
\[
\tilde{y_i} =
\begin{cases} 
y_i^{O} & \text{the sample was not converted before testing}, \\
y_i^{T} & \text{the sample converted before testing}. 
\end{cases}
\]
That is, retrain loss assumes access to all true conversions observed up to the evaluation time, representing the best achievable practical performance.
\subsection{Influence Function} 
Influence function \cite{hampel1974influence}, which is a classical tool in robust statistics for measuring the changes of the model parameters to small changes in the weights of training samples, has been recently introduced in the machine learning community for understanding black-box predictions and beyond \cite{koh2017understanding}. Given a training data set $\mathcal{Z} = \{z_1, z_2, ..., z_n\}$, we consider the following empirical risk minimization problem:
\begin{equation}
\hat{\theta} \in \mathop{\arg\min}_{\theta} R(\theta) := \frac{1}{n} \sum_{i=1}^n \mathcal{L}\left(z_i, \theta\right),
\end{equation}
where $\mathcal{L}(\cdot, \theta)$ is the loss function (\eg Equations \eqref{vanilla} \eqref{oracle}), $\theta$ is the model parameter.

We are interested in estimating the model parameter change $\Delta\theta(\epsilon) = \hat{\theta}_{{new}}(\epsilon) - \hat{\theta}$ if some training sample $z_j$ is slightly reweighted by $\epsilon$, where
\begin{equation}
\label{eq: if-newtheta}
\hat{\theta}_{{new}}(\epsilon) \in \mathop{\arg\min}_{\theta} ~ \widehat{\mathcal{L}}(z_j; \theta, \epsilon) = R(\theta) + \epsilon\mathcal{L}(z_j, \theta).
\end{equation}
When $\epsilon \approx 0$, the influence function provides an elegant tool for estimating $\Delta\theta(\epsilon)$ without solving \eqref{eq: if-newtheta} as
\begin{align}
\label{dalta}
\Delta \theta(\epsilon) \approx -\epsilon H_{\hat{\theta}}^{-1} \nabla_\theta \mathcal{L}(z_j, \hat{\theta}) \notag \\~\mbox{and}~ \left.\frac{d \Delta\theta(\epsilon)}{d \epsilon}\right|_{\epsilon=0} = -H_{\hat{\theta}}^{-1} \nabla_\theta \mathcal{L}(z_j, \hat{\theta}),
\end{align}
where $H_{\hat{\theta}} = \nabla_\theta^2 R(\hat{\theta})$ is the Hessian matrix of $R(\cdot)$ at $\theta = \hat{\theta}$. A detailed derivation of the expression in \eqref{dalta} can be found in \cite{koh2017understanding}. 

The key to applying the influence function for estimating $\Delta\theta$ lies in efficiently computing the right-hand side of \eqref{dalta}, which is challenging in practice due to high-dimensional parameters and large-scale data. As a core contribution, we propose a novel stochastic algorithm to tackle this computational bottleneck and unlock the practical potential of influence functions.

\section{Method}
Delayed feedback in CVR prediction presents challenges from both label reversal and new data integration. We propose IF-DFM, a unified framework that addresses both issues without retraining, and introduce a stochastic algorithm to reduce the computational cost of influence function estimation.
\subsection{IF-DFM}
In this section, we elaborate on the proposed IF-DFM framework, which models both label reversals and newly arrived data as data perturbations. 
\subsubsection{Label Reversal}
In CVR prediction, delays between click and conversion behaviors often lead to label reversals, where some fake negatives become positives after training. This phenomenon can be naturally modeled as data perturbation. Specifically, if a label reversal occurs for a training sample $z_i$, it can be represented as perturbing $z_i$ to $z_i^\delta = (x_i, y_i^{O} + \delta)$. Since observed positives are correctly labeled, the perturbation for CVR is $\delta = 1$.
Without loss of generality, we consider the case where a label reversal occurs for $z_i$, and study the perturbed empirical risk minimization by reweighting this sample:
\begin{equation}
\small
\label{eq: perturbed-ER}
\widehat{\theta_\delta}(\epsilon) \in \mathop{\arg\min}_{\theta} \left( \mathcal{L}_{{V}}(\theta)+\epsilon \mathcal{L}_{{BCE}}\left(z_i^{\delta},\theta\right)-\epsilon  \mathcal{L}_{{BCE}}\left(z_i, \theta\right) \right).
\end{equation}
Note that the right-hand side of \eqref{eq: perturbed-ER} coincides with the retrain loss \eqref{retrain} if we take $\epsilon = 1/n$. When $\epsilon \approx 0$ (which is true if $\epsilon = 1/n$ and $n$ is sufficiently large), following the derivation of influence function (\eg see \cite{koh2017understanding}), we can obtain that  
\begin{equation}
\label{eq: delta-estimate-delay}
\begin{aligned}
\left.\frac{d \widehat{\theta_\delta}}{d \epsilon}\right|_{\epsilon=0} &=
 -\left(\nabla_\theta^2 \mathcal{L}_{V} (\hat{\theta}) \right)^{-1} \big( \nabla_\theta \mathcal{L}_{\text{BCE}}\left(z^\delta_i, \hat{\theta}\right) \\
&\quad - \nabla_\theta \mathcal{L}_{\text{BCE}}(z_i, \hat{\theta}) \big),
\end{aligned}
\end{equation}
where $\hat{\theta}$ is a minimizer of \eqref{vanilla} (which is available if the Vanilla model is trained).

Therefore, after aggregating the gradients of all label-reversed samples, we can estimate the changes of the model parameter from \eqref{eq: delta-estimate-delay} by taking $\epsilon = 1/n$ as
\begin{equation}
\small
\label{eqc}
\begin{aligned}
\Delta \theta_{\text{delay}} \approx &\frac{1}{n}\left(\nabla_\theta^2 \mathcal{L}_{V}(\hat{\theta})\right)^{-1} \bigg(
\sum_{j\in J} \nabla_\theta \mathcal{L}_{\text{BCE}}\left(z_j, \hat{\theta}\right) \\
&- \sum_{j \in J} \nabla_\theta \mathcal{L}_{\text{BCE}}\left(z_j^\delta, \hat{\theta}\right) \bigg),
\end{aligned}
\end{equation}
where $J$ is the index set of samples that clicked before $T$ and converted during the $[T, T']$ interval (\ie samples with label reversal). 

\subsubsection{Newly Arrived Behavioral Data Integration}
Conventionally, the influence functions estimate how model parameters or test loss change when a sample is removed or altered. In our IF-DFM framework, we extend this to incorporate new behavioral data arriving during the interval $[T, T']$. To this end, we consider the following perturbed loss function for a new sample $z = (x, y)$ and $0 \leq \epsilon < 1$:
\begin{equation}
\label{eq: perturbed-loss-add}
\widehat{\theta_z}(\epsilon) \in \mathop{\arg\min}_{\theta} \left((1 - \epsilon)\mathcal{L}_{{V}}(\theta)+\epsilon \mathcal{L}_{{BCE}}\left(z,\theta\right)\right).
\end{equation}
Note that we can recover the retrain loss for integrating the additional new sample $z$ by taking $\epsilon = 1/(n+1)$. By careful calculations (which can be found in Appendix), we can find an approximation to $\Delta_{z, \epsilon} = \widehat{\theta_z}(\epsilon) - \hat{\theta}$ for sufficiently small $\epsilon > 0$ as
\begin{equation}
\label{eq: Delta_incremental}
    \Delta_{z, \epsilon} \approx -\frac{\epsilon}{1 - \epsilon}\left(\nabla^2_{\theta}\mathcal{L}_V(\hat{\theta})\right)^{-1}\nabla_{\theta}\mathcal{L}_{BCE}(z, \hat{\theta}).
\end{equation}
Therefore, for sufficiently large $n$, we obtain an approximation to the change of model parameter for integrating the new data sample $z$ as
\begin{align}
\label{delta-add}
    \Delta\theta_{add} &= \widehat{\theta_z}\left(\frac{1}{n+1}\right) - \hat{\theta} \notag \\ & \approx -\frac{1}{n}\left(\nabla^2_{\theta}\mathcal{L}_V(\hat{\theta})\right)^{-1}\nabla_{\theta}\mathcal{L}_{BCE}(z, \hat{\theta}).
\end{align}

Considering both the label reversed data and the newly arrived data, the equation for the total parameter change can be given as
\begin{equation}
\Delta \theta_{total} = \Delta \theta_{delay} + \Delta \theta_{add}.
\end{equation}
More specifically, we need to estimate 
\begin{align}
\label{total}
\Delta &\theta_{\text{total}}  \approx \left(\nabla_\theta^2 \mathcal{L}_{V}(\hat{\theta})\right)^{-1} \bigg(\frac{1}{n} \sum_{j \in J} \nabla_\theta \mathcal{L}_{V}(z_j, \hat{\theta}) \notag \\
&\quad - \frac{1}{n} \sum_{j \in J} \nabla_\theta \mathcal{L}_{V}(z_j^\delta, \hat{\theta})- \frac{1}{n} \sum_{k \in K} \nabla_\theta \mathcal{L}_{V}(z_k, \hat{\theta}) \bigg),
\end{align}
where $J$ and $K$ are the index sets for the label reversed data samples and newly arrived data samples, respectively.

Addressing problem \eqref{total} poses significant computational challenges. When the Hessian matrix $\nabla_{\theta}^2\mathcal{L}_{V}(\hat{\theta})$ is positive definite, we can apply the conjugate gradient (CG) method to estimate $\Delta \theta_{total}$, where only Hessian-vector products are required. For $\theta \in \mathbb{R}^p$ (where $p$ is usually huge in applications), the CG method requires at most $p$ iterations to calculate the right-hand side of \eqref{total} and it usually requires much fewer iterations to obtain an approximate solution if the condition number of $\nabla_{\theta}^2\mathcal{L}_{V}(\hat{\theta})$ is close to $1$ \cite{nocedal1999numerical}. However, the CG method typically requires full-batch gradient and Hessian matrix computations, making it impractical for large-scale datasets. To address the computational challenges of the CG method, an approximation of $\left(\nabla^2_{\theta}\mathcal{L}_V(\hat{\theta})\right)^{-1}$ via its truncated power series has been widely used to estimate $\Delta\theta_{total}$ \cite{koh2017understanding}. However, its accuracy is often compromised. Finding a more precise and efficient solution for \eqref{total} remains a challenge. Next, we will address this challenge by designing an efficient and scalable stochastic algorithm.

\subsection{An Efficient and Scalable Method for Calculating the parameter changes $\Delta\theta_{total}$}
In this section, we will address the computational challenges for calculating $\Delta\theta_{total}$. Our core idea is to compute $\Delta\theta_{total}$ by converting $\eqref{total}$ into an equivalent finite-sum optimization problem. Consequently, we can apply popular scalable algorithms, such as the stochastic gradient descent \cite{robbins1951stochastic} and its variants \cite{duchi2011adaptive,kingma2014adam,johnson2013accelerating}, to calculate $\Delta\theta_{total}$ efficiently. The definition of $\Delta\theta_{total}$ implies that it is a solution to the linear system 
\begin{equation}
\label{eq: Equiv-ls}
    \nabla^2_{\theta}\mathcal{L}_V(\hat{\theta}) \Delta = b
\end{equation}
with
\begin{align}
\label{def-b}
\small
b = &\frac{1}{n}\sum_{j \in J} \nabla_\theta \mathcal{L}_{{V}}\left(z_j, \hat{\theta}\right) - \frac{1}{n}\sum_{j \in J} \nabla_\theta \mathcal{L}_{{V}}\left(z_j^\delta, \hat{\theta}\right) \\ & - \frac{1}{n}\sum_{k \in K} \nabla_\theta \mathcal{L}_{{V}}\left(z_k, \hat{\theta}\right).
\end{align}
Since $\hat{\theta}$ is a minimizer of \eqref{vanilla}, it satisfies the second-order necessary optimality condition~\cite{nocedal1999numerical}, implying that the Hessian $\nabla^2_{\theta}\mathcal{L}_V(\hat{\theta})$ is symmetric and positive semidefinite. As a result, $\Delta\theta_{total}$ corresponds to the solution of a convex quadratic optimization problem:

\begin{equation}
\mathop{\min}_{\Delta} ~ F(\Delta) := \frac{1}{2}\Delta^{\top} \nabla^2_{\theta}\mathcal{L}_V(\hat{\theta}) \Delta - \langle b, \Delta \rangle.
\end{equation}
It follows from the finite-sum structure of $\mathcal{L}_V$ that
\begin{equation}
\label{eq: finite-sum}
    F(\Delta) = \frac{1}{n}\sum_{i = 1}^n f_i(\Delta),
\end{equation}
where
\begin{equation}
\label{eq: fun-fi}
f_i(\Delta) = \frac{1}{2} \Delta^{\top} \nabla_{\theta}^2\mathcal{L}_{BCE}(z_i, \hat{\theta}) \Delta - \langle b, \Delta \rangle.
\end{equation}
Therefore, we have converted the calculation of $\Delta\theta_{total}$ equivalently to minimizing an optimization problem with a finite-sum objective function. Note that the function value and the gradient of $f_i$ can be efficiently evaluated via the built-in auto-differentiation and the Hessian vector product modules implemented by popular frameworks, such as PyTorch. Consequently, we can apply efficient and scalable optimization algorithms to calculate $\Delta\theta_{total}$. In particular, we choose ADAM \cite{kingma2014adam} in this paper. We summarize the details in Algorithm \ref{alg1}.

\begin{algorithm}[htbp]
\caption{An efficient algorithm to calculate $\Delta\theta_{total}$.}\label{alg1}
\begin{algorithmic}[1]
\STATE {\textbf{Input:}} Training data $\mathcal{D}$; Label reversed data $z_j \rightarrow z_j^{\delta}$; Newly arrived data $z_k$; Trained vanilla model $\hat{\theta}$.
\STATE{\textbf{Output:}} The model parameter change $\Delta\theta_{total}$.

\STATE{\textbf{Initialization:}} Construct $b$ in \eqref{def-b} and the function $f_i$.
\STATE Call ADAM to minimize \eqref{eq: finite-sum} and obtain $\Delta^*$.
 \STATE Return $\Delta\theta_{total} = \Delta^*$.
\end{algorithmic}
\end{algorithm}
\section{Experiment}
Our experiments aim to investigate the following research questions: \textbf{RQ1}: How does our method's performance compare to state-of-the-art methods in CVR prediction? \textbf{RQ2}: Can our method adapt to dynamic changes in user interests over time? \textbf{RQ3}: How efficient is our method in calculating parameter changes?

\subsection{Experimental Settings}
\subsubsection{Datasets.}
To assess our method, we conduct experiments on two large-scale datasets: Criteo and Taobao. The statistics of these datasets are summarized in Table \ref{dataset}.
\begin{table}[h]
\small
\begin{center}
\setlength{\tabcolsep}{5pt}
\caption{Statistics of Criteo and Taobao datasets.}
\scriptsize
\label{dataset}
\begin{tabular}{ccc|ccc}
\toprule
{Name}    & {\# clicks} & {CVR} & {Name}   & {\# clicks} & {CVR} \\ \midrule
{Criteo}   &    4,019,339       &  0.2227   & {Taobao}     &   8,544,800        &   0.0100  \\ \bottomrule
\end{tabular}
\end{center}
\end{table}

\textbf{Criteo}\footnote{\url{https://labs.criteo.com/2013/12/conversion-logs-dataset/}}: A public dataset with click logs from Criteo's live traffic, containing timestamps, conversion labels, and both categorical and continuous features. We use 3 million samples spanning 14 days for training.

\textbf{Taobao}\footnote{\url{https://tianchi.aliyun.com/dataset/dataDetail?dataId=649&userId=1&lang=en-us}}: A dataset of user interactions on Taobao, including behaviors such as page views and purchases, along with timestamps and associated features.

Following common settings in previous offline research \cite{wang2023unbiased,yasui2020feedback}, we employ a temporal partitioning approach to divide each dataset into training, validation, and test sets. Specifically, we use data from the interval $\left[T^{\prime}-d_\text{test}, T^{\prime}\right]$ for validation and data from $\left[T^{\prime}, T^{\prime}+d_\text{test}\right]$ for testing, where $d_\text{test}$ represents the size of the validation/test set.  The interval $\left[T, T^{\prime}\right]$ covers a period of $c$ days. We will explore the value of $c$ in the following sections. Such a setting enables us to assess the CVR model's ability to capture evolving user interests.

\subsubsection{Metrics.}
We adopt three standard metrics to evaluate CVR performance following \cite{wang2023unbiased, dai2023dually}: \textbf{AUC} (Area Under the ROC Curve), which measures the ranking quality between positive and negative samples; \textbf{PRAUC} (Precision-Recall AUC), which evaluates performance across precision-recall thresholds; and \textbf{Log Loss (LL)}, which assesses prediction accuracy via probabilistic error (lower is better). We also report the \textbf{Relative Improvement (RI)} over the Vanilla model, normalized by the performance gap between Retrain and Vanilla. For a method $f$ and metric $M$, RI is computed as:
\begin{equation}
\mathrm{RI}_M = \frac{M(f) - M(\text{Vanilla})}{M(\text{Retrain}) - M(\text{Vanilla})}.
\end{equation}
This yields RI-AUC, RI-PRAUC, and RI-LL, reflecting how much each method narrows the gap to Retrain.

\subsubsection{Baselines.}
We compare our method with 13 state-of-the-art baselines, covering both offline (\eg Vanilla, Retrain, DFM \cite{chapelle2014modeling}, FSIW \cite{yasui2020feedback}, nnDF \cite{kato2020learning}, ULC \cite{wang2023unbiased}) and online approaches (e.g., Pretrain, Retrain-online, FNC \cite{ktena2019addressing}, FNW \cite{ktena2019addressing}, DDFM \cite{dai2023dually}, ES-DFM \cite{yang2021capturing}, DEFUSE \cite{chen2022asymptotically}). For fair comparison, we follow \cite{wang2023unbiased} and adopt four widely-used CVR backbones: MLP, DeepFM \cite{guo2017deepfm}, AutoInt \cite{song2019autoint}, and DCNV2 \cite{wang2021dcn}. Detailed descriptions of all baselines are provided in Appendix.

\subsubsection{Implementation Details.}
In the offline setting, IF-DFM leverages the influence function to estimate $\Delta\theta_{delay}$ and adjust the Vanilla model's parameters. In the online setting, we use the offline Vanilla model as the pretrained model. For methods requiring auxiliary models or variables (\eg DDFM), we follow their pretraining methods. The model parameters are modified based on the approximate $\Delta\theta_{total}$. More implementation details can be found in Appendix.

\subsection{Performance Comparison}
\begin{table*}[t]
\small
\centering
\setlength{\tabcolsep}{10pt} 
\caption{Experimental results on Criteo. The superscripts ** indicate $p \leq 0.05$ for the t-test of IF-DFM vs. the best baseline.}
\label{main}
\scriptsize
\begin{tabular}{c|c|cccccc}
\toprule
{Backbone}                 & {Method}  & \multicolumn{1}{c}{{AUC ↑}} & \multicolumn{1}{c}{{PRAUC ↑}} & \multicolumn{1}{c}{{LL ↓}} &\multicolumn{1}{c}{{RI-AUC ↑}} & \multicolumn{1}{c}{{RI-PRAUC ↑}} & {RI-LL ↑}         \\ \midrule
\multirow{7}{*}{{MLP}}     & {Vanilla} & 0.8353                     & 0.6398                       & 0.4187                    & 0.000                         & 0.000                           & 0.000           \\
                         & {Retrain} & 0.8419                     & 0.6513                       & 0.3930                    & 1.000                         & 1.000                           & 1.000           \\ \cline{2-8} 
                         &  {DFM}     & 0.8355                     & 0.6409                       & 0.4152                    & 0.0303                        & 0.0957                          & 0.1362          \\
                         & {FSIW}    & 0.8369                     & 0.6433                       & 0.4020                    & 0.2424                        & 0.3043                          & 0.6498          \\
                         & {nnDF}    & 0.6887                     & 0.4029                       & 0.5639                    & -22.21                        & -20.60                          & -5.670          \\
                         & {ULC}     & 0.8343                     & 0.6412                       & 0.3994                    & -0.1515                       & 0.1217                          & 0.7510          \\
                         & \textbf{IF-DFM (Ours)}   &  \textbf{0.8411**}               & \textbf{0.6491**}               & \textbf{0.3958**}           & \textbf{0.8788**}               & \textbf{0.8087**}                 & \textbf{0.9662**} \\ \midrule
\multirow{7}{*}{{DeepFM}} & {Vanilla} & 0.8368                     & 0.6411                       & 0.4168                    & 0.000                         & 0.000                           & 0.000           \\
                         & {Retrain} & 0.8418                     & 0.6509                       & 0.3926                    & 1.000                         & 1.000                           & 1.000           \\ \cline{2-8} 
                         & {DFM}     & 0.8381                     & 0.6434                       & 0.4065                    & 0.260                         & 0.2347                          & 0.4256          \\
                         & {FSIW}    & 0.8378                     & 0.6436                       & 0.4023                    & 0.200                         & 0.2551                          & 0.5992          \\
                         & {nnDF}    & 0.7025                     & 0.4146                       & 0.5651                    & -26.86                        & -23.11                          & -6.128          \\
                         & {ULC}     & 0.8376                     & 0.6446                       & 0.3982                    & 0.160                         & 0.3571                          & 0.7686          \\
                         & \textbf{IF-DFM (Ours)}    & \textbf{0.8412**}            & \textbf{0.6492**}              & \textbf{0.3962**}           & \textbf{0.8800**}               & \textbf{0.8265**}                 & \textbf{0.8512**} \\ \midrule
\multirow{7}{*}{{AutoInt}} & {Vanilla} & 0.8361                     & 0.6401                       & 0.4174                    & 0.000                         & 0.000                           & 0.000           \\
                         & {Retrain} & 0.8415                     & 0.6503                       & 0.3929                    & 1.000                         & 1.000                           & 1.000           \\ \cline{2-8} 
                         & {DFM}     & 0.8374                     & 0.6426                       & 0.4091                    & 0.2407                        & 0.2451                          & 0.3388          \\
                         & {FSIW}    & 0.8376                     & 0.6442                       & 0.4026                    & 0.2778                        & 0.4020                          & 0.6041          \\
                         & {nnDF}    & 0.6820                     & 0.3770                       & 0.5789                    & -28.54                        & -25.79                          & -6.592          \\
                         & {ULC}     & 0.8374                     & 0.6443                       & 0.3987                    & 0.2407                        & 0.4118                          & 0.7633          \\
                         & \textbf{IF-DFM (Ours)}    & \textbf{0.8404**}            & \textbf{0.6475**}              & \textbf{0.3965**}           & \textbf{0.7963**}               & \textbf{0.7255**}                 & \textbf{0.8530**} \\ \midrule
\multirow{7}{*}{{DCNV2}}  & {Vanilla} & 0.8370                     & 0.6418                       & 0.4174                    & 0.000                         & 0.000                           & 0.000           \\
                         & {Retrain} & 0.8417                     & 0.6508                       & 0.3927                    & 1.000                         & 1.000                           & 1.000           \\ \cline{2-8} 
                         & {DFM}     & 0.8372                     & 0.6422                       & 0.4087                    & 0.0426                        & 0.044                           & 0.3522          \\
                         & {FSIW}    & 0.8388                     & 0.6456                       & 0.3997                    & 0.3830                        & 0.422                           & 0.6045          \\
                         & {nnDF}    & 0.6850                     & 0.3878                       & 0.5728                    & -32.34                        & -28.22                          & -6.291          \\
                         & {ULC}     & 0.8368                     & 0.6441                       & 0.3989                    & -0.0426                       & 0.2556                          & 0.7490          \\
                         & \textbf{IF-DFM (Ours)}    & \textbf{0.8410**}            & \textbf{0.6501**}              & \textbf{0.3985**}           & \textbf{0.8511**}               & \textbf{0.9222**}                 & \textbf{0.7652**}  \\ \bottomrule
\end{tabular}
\end{table*}

\begin{figure}[t]
  \centering
  \begin{minipage}[t]{0.225\textwidth}
    \centering
\includegraphics[width=\textwidth]{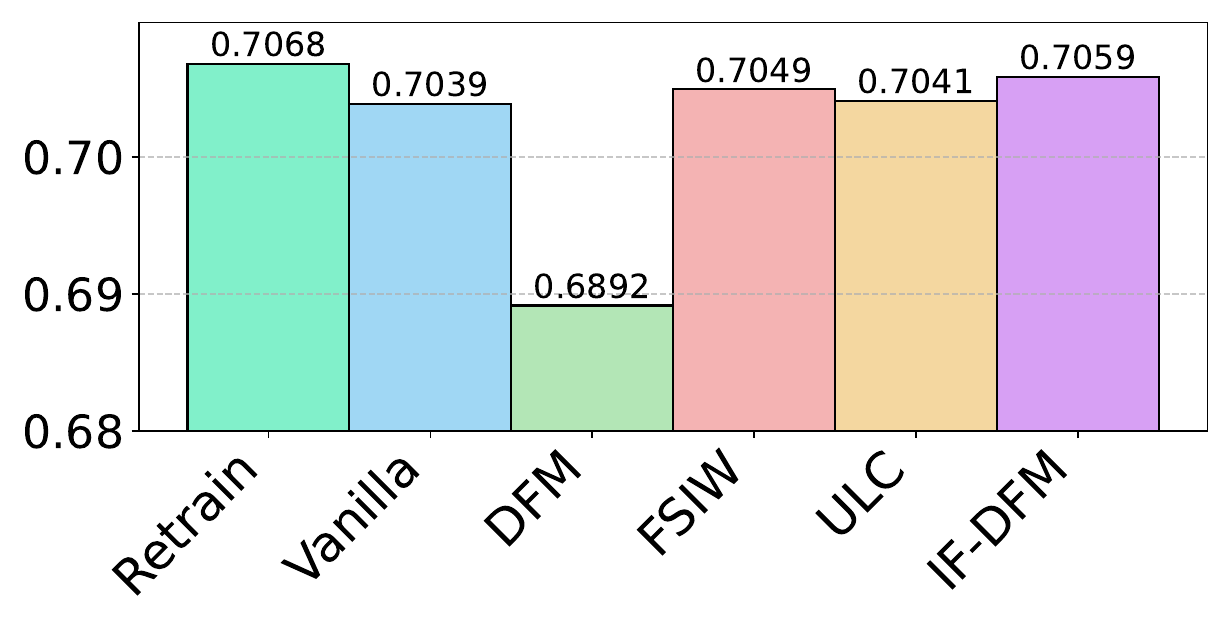}
    \caption*{(a) AUC Results}
  \end{minipage}
  \begin{minipage}[t]{0.225\textwidth}
    \centering
\includegraphics[width=\textwidth]{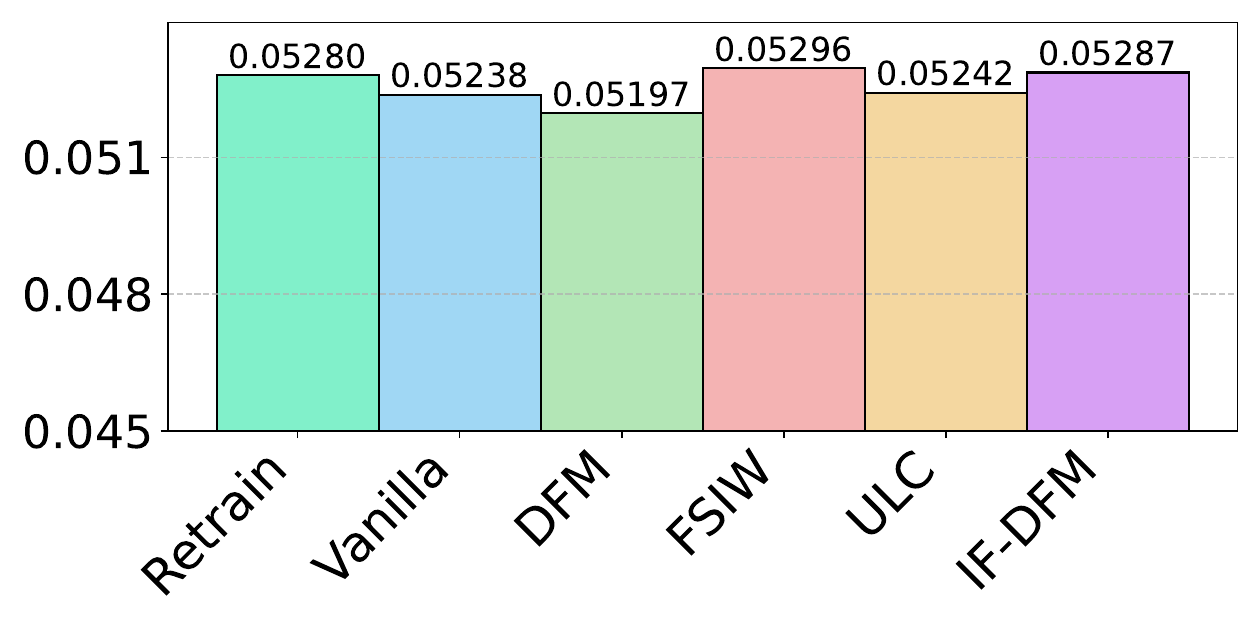}
    \caption*{(b) PRAUC Results}
  \end{minipage}
  \caption{Offline experimental results on Taobao dataset.}
  \label{fig:both_images}
\end{figure}

The offline results on the Criteo dataset across different backbones are summarized in Table~\ref{main}, from which we draw the following observations:
\begin{itemize}[leftmargin=*]
\item Retrain consistently outperforms Vanilla, confirming that delayed feedback negatively impacts model performance. Both FSIW and DFM improve upon Vanilla, with FSIW achieving better results due to its pre-trained auxiliary models. In contrast, nnDF underperforms due to its reliance on global information and the need for full data during updates.
\item ULC shows limited performance, as it depends on a label correction model that requires high-quality historical data and struggles to adapt to shifting user interests.
\item IF-DFM significantly outperforms all baselines across metrics and backbones, demonstrating its effectiveness in handling delayed feedback. It achieves average improvements of $0.55\%$ in AUC, $1.29\%$ in PRAUC, and $4.99\%$ in LL. Prior work from Google~\cite{cheng2016wide,wang2017deep} and Microsoft~\cite{ling2017model} indicates that even a $0.1\%$ AUC gain can lead to substantial real-world benefits.
\item In terms of relative gains, IF-DFM approaches the performance of Retrain, with improvements of $85.16\%$ (AUC), $82.07\%$ (PRAUC), and $85.89\%$ (LL), underscoring its effectiveness.
\end{itemize}
Compared to Criteo, the Taobao dataset has a lower conversion rate. As shown in Figure~\ref{fig:both_images}, IF-DFM surpasses the best baseline in most cases, with AUC and PRAUC gains of $0.28\%$ and $0.94\%$, respectively. FSIW benefits from pre-trained importance estimators, giving it short-term advantages. However, IF-DFM, by leveraging real label information, better approximates retraining and, as shown in subsequent experiments, adapts more effectively to evolving user interests.

\begin{figure}[h]
  \centering
  \begin{minipage}[t]{0.230\textwidth}
    \centering
\includegraphics[width=\textwidth]{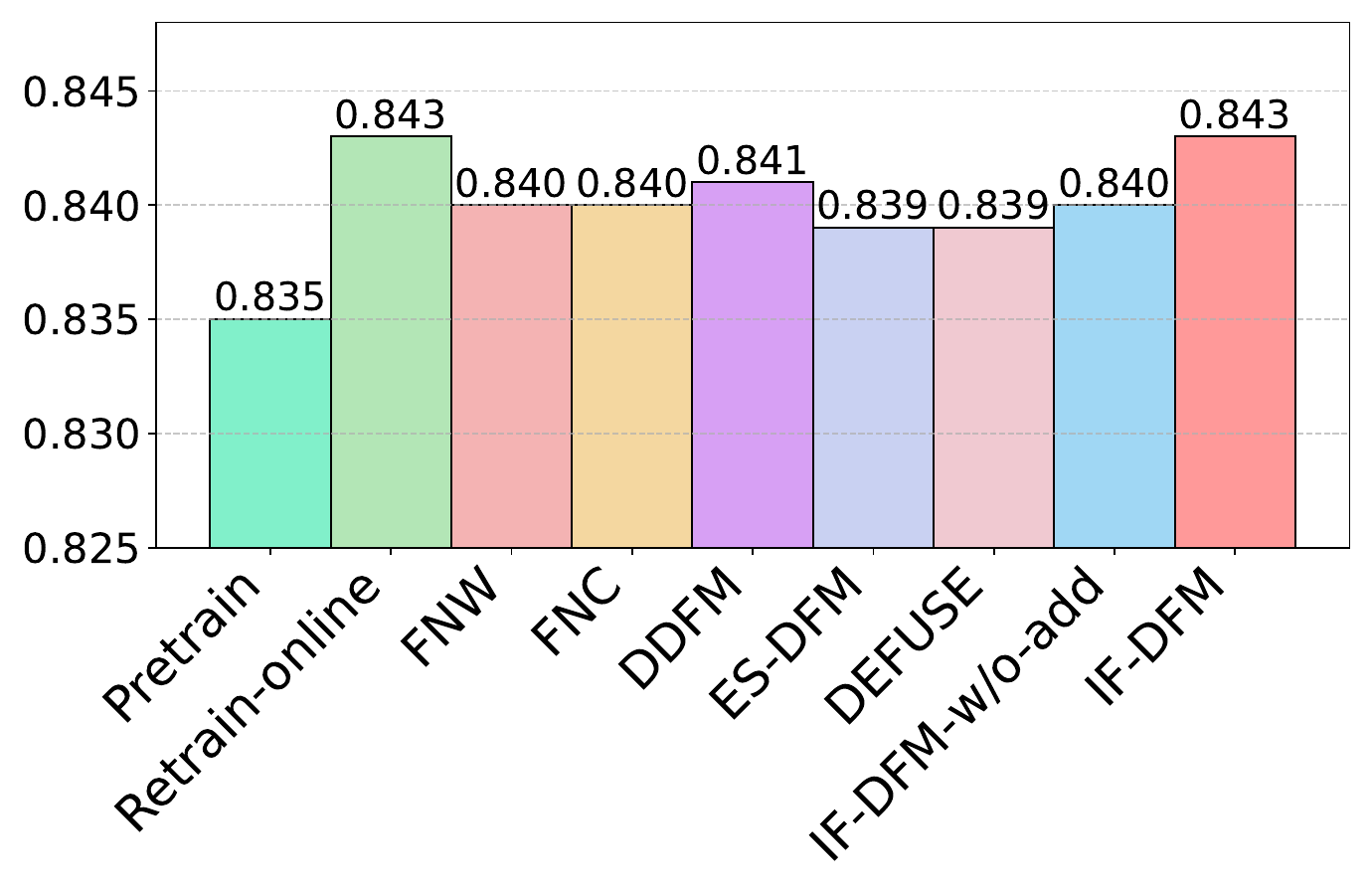}
    \caption*{(a) AUC Results}
  \end{minipage}
  \begin{minipage}[t]{0.230\textwidth}
    \centering
    \includegraphics[width=\textwidth]{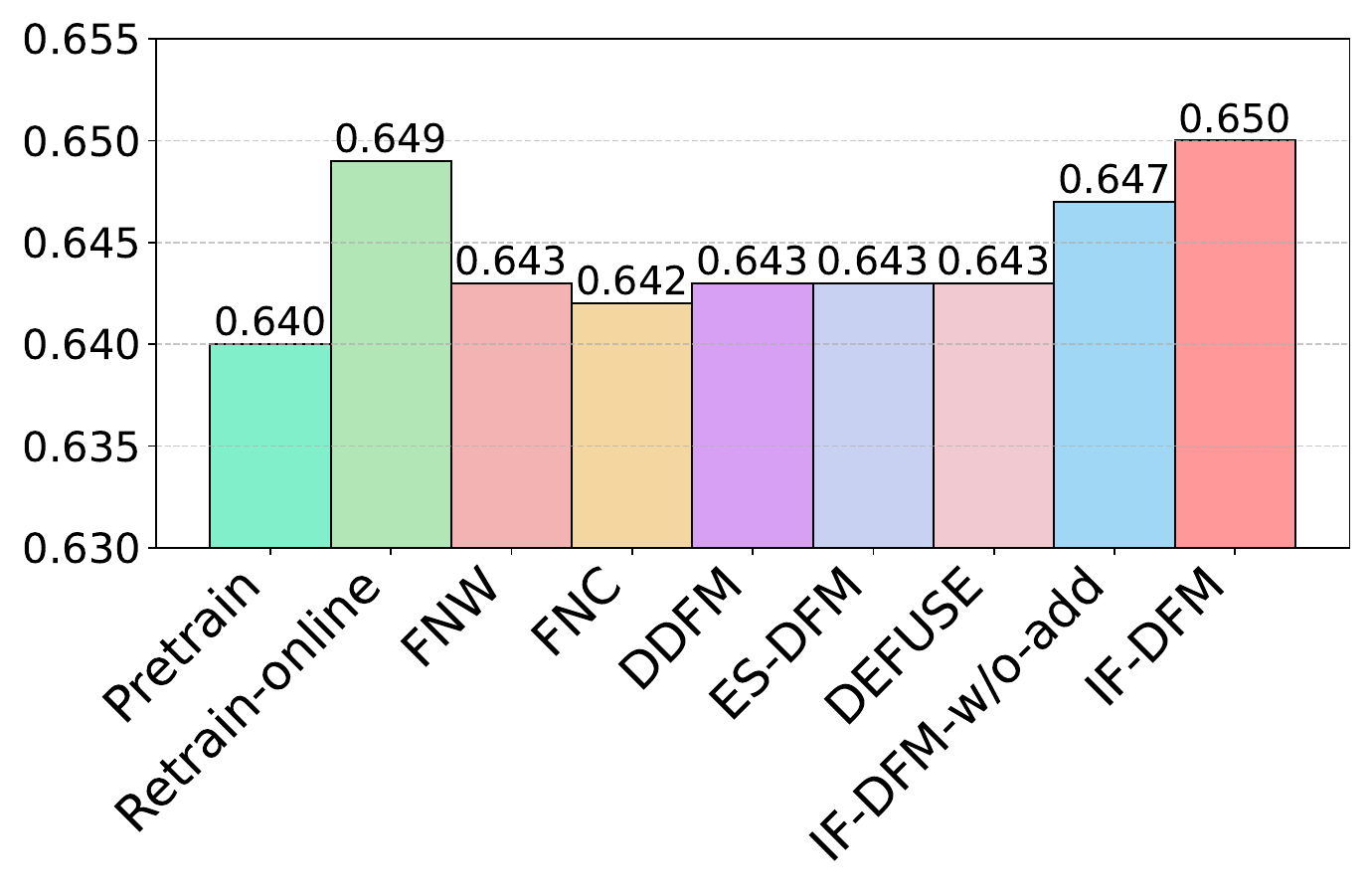}
    \caption*{(b) PRAUC Results}
  \end{minipage}
  \caption{Online experimental results on Criteo dataset.}
  \label{fig:incre}
\end{figure}

\subsection{Effectiveness of Adapting to Evolving User Interest}
User interest drift is a critical challenge in CVR prediction, as outdated models may fail to capture evolving preferences, leading to degraded performance. Table~\ref{table:c} reports the results of various methods under different values of $c$, which denotes the time gap between training and test sets. More results can be found in Appendix. From these results, we can draw the following conclusions:
\begin{itemize}[leftmargin=*]
\item ULC and DFM degrade notably over time. At $c=14$, both perform worse than Vanilla. FSIW shows some resilience by adaptively reweighting samples based on feedback, but its effectiveness is limited as the temporal gap grows. ULC, constrained by its use of counterfactual deadlines, fails to generalize well.
\item The performance gap between IF-DFM and baselines grows with increasing $c$, highlighting its robustness to temporal drift. By estimating the influence of label reversed behavior data, IF-DFM effectively approximates retraining without incurring its full cost.

\end{itemize}

\begin{table}[t]
\small
\centering
\caption{Offline experimental results on the Criteo dataset for different values of $c$, with MLP as the backbone.}
\begin{tabular}{cc|ccc}
\toprule
    \multicolumn{2}{c|}{\textbf{}}                                            & {AUC ↑}                & {PRAUC ↑}              & {LL ↓}                 \\ \midrule
    \multicolumn{1}{c|}{}                                  & {Vanilla} & 0.8427                      & 0.6479                      & 0.4085                      \\
    \multicolumn{1}{c|}{}                                  & {Retrain} & {0.8464} & {0.6541} & {0.3879} \\
     \cline{2-5}
    \multicolumn{1}{c|}{}                                  & {nnDF}    & 0.6840                        & 0.3979                        & 0.5565                        \\
    \multicolumn{1}{c|}{}                                  & {DFM}     & 0.8412                        & 0.6469                        & 0.4064                        \\
    \multicolumn{1}{c|}{}                                  & {FSIW}    & 0.8441                        & 0.6508                        & 0.3898                        \\
    \multicolumn{1}{c|}{}                                  & {ULC}     & 0.8445                        & 0.6513                        & 0.3894                      \\
    \multicolumn{1}{c|}{\multirow{-7}{*}{{$c$ = 5}}}  & \textbf{Ours}    & \textbf{0.8452}               & \textbf{0.6514}               & 
    \textbf{0.3883}             \\  \midrule
    \multicolumn{1}{c|}{}                                  & {Vanilla} & 0.8369                      & 0.6385                      & 0.4114                      \\ 
    \multicolumn{1}{c|}{}                                  & {Retrain} &  0.8453 &  0.6529 &  0.3860 \\\cline{2-5}
    \multicolumn{1}{c|}{}                                  & {nnDF}    & 0.7041                        & 0.4116                        & 0.5521                        \\
    \multicolumn{1}{c|}{}                                  & {DFM}     & 0.8351                        & 0.6370                        & 0.4076                        \\
    \multicolumn{1}{c|}{}                                  & {FSIW}    & 0.8385                        & 0.6421                        & 0.3965                        \\
    \multicolumn{1}{c|}{}                                  & {ULC}     & 0.8345                        & 0.6385                        & 0.3954                        \\
    \multicolumn{1}{c|}{\multirow{-7}{*}{{$c$ = 14}}} & \textbf{Ours}    & \textbf{0.8420}               & \textbf{0.6463}               & \textbf{0.3938} \\\bottomrule
\end{tabular}
\label{table:c}
\end{table}

In the online setting, IF-DFM estimates parameter updates from both fake negatives and newly arrived data. We also evaluate a variant, IF-DFM-w/o-add, which omits updates from new data ($\Delta{\theta_{add}}$), using only delayed conversions ($\Delta{\theta_{delay}}$). The results are shown in Figure \ref{fig:incre}. We can find that:

\begin{itemize}[leftmargin=*]
    \item Compared to pretraining, online CVR models can mitigate the delayed feedback problem to some extent. This can be attributed to their fine-tuning pipeline using new data and delayed conversion data.
    \item IF-DFM outperforms IF-DFM-w/o-add, with AUC and PRAUC gains of $0.36\%$ and $0.47\%$, respectively, indicating the benefit of incorporating newly arrived data.
    \item IF-DFM exceeds all baselines and closely approaches retraining performance, demonstrating its ability to effectively integrate delayed and new label information for improved adaptability.
\end{itemize}

\subsection{Efficiency of Calculating Parameter Changes}
Ensuring rapid updates of model parameters is crucial for maintaining the freshness of a CVR model. Due to space constraints, we report runtime results only for methods with the MLP backbone on the Criteo dataset, as shown in Figure~\ref{time}. We have the following observations:
\begin{itemize}[leftmargin=*]
    \item Compared to Vanilla, ULC requires the training of additional auxiliary models and the alternating training of two models, nearly doubling the training time. This indicates that relying on auxiliary models incurs extra time costs, making it inefficient and computationally expensive.
    \item IF-DFM updates model parameters without retraining, taking only 14.8 seconds --- just $1.1\%$ of Vanilla’s training time --- demonstrating its high efficiency in addressing delayed feedback.

\end{itemize}

\begin{table}[t]
\caption{Running time of different methods under the offline setting, with MLP as the backbone and $c=10$.}
\centering
\scriptsize
\begin{tabular}{c|ccccc}
\toprule 
     & 
     {Vanilla} & {ULC}       & {DFM}     & {FSIW}    & {Vanilla + Ours}  \\ \midrule
\textbf{Time (s)} & 1351.4 & 2467.4 & 1979.8 & 1018.5 & 1351.4 + 14.8 \\ \bottomrule
\end{tabular}
\label{time}
\end{table}

\section{Related work}
\subsection{Delayed Feedback}
Delayed feedback is a central issue in CVR modeling. Offline methods assume i.i.d. data, using historical samples to predict future conversions. DFM~\cite{chapelle2014modeling} models delay with an exponential distribution, while NoDeF~\cite{yoshikawa2018nonparametric} adopts a non-parametric approach. Methods like FSIW~\cite{yasui2020feedback}, nnDF~\cite{kato2020learning}, and ULC \cite{wang2023unbiased} aim to construct unbiased estimators of oracle loss through reweighting or label correction. However, most require training auxiliary models, hindering adaptability to evolving user interests. Online methods update models with real-time feedback. ES-DFM~\cite{yang2021capturing} uses elapsed-time sampling for delayed positives, DEFUSE~\cite{chen2022asymptotically} employs a bi-distribution strategy, and DDFM~\cite{dai2023dually} uses dual estimators for different data phases. Yet, duplicating delayed conversions may introduce ambiguity, and many approaches still depend on auxiliary components.

\subsection{Influence Function}
Influence functions estimate the effect of individual training points on model outputs. Introduced by Koh \etal \cite{koh2017understanding}, they have since been applied to interpretability \cite{chhabra2024data,basu2020second}, fairness \cite{li2022achieving,wang2024fairif}, data management \cite{hara2019data}, unlearning \cite{chen2024fast, liu2023muter}, and recommendation systems \cite{cheng2018explaining, guo2024counterfactual,li2023selective}. Extensions include robustness analysis \cite{koh2019accuracy} and data debugging via TracIn \cite{pruthi2020estimating}. IF4URec \cite{yu2020influence} improves recommender systems through sample reweighting. However, as Basu \etal \cite{basu2020influence} highlight, scalability remains a challenge in deep learning, and although FastIF \cite{guo2020fastif} improves efficiency to some extend, it does not touch the core of the inverse Hessian-vector product computation, limiting applicability to large models.
\section{Conclusion}
We propose IF-DFM, an influence function-based framework for modeling delayed feedback. IF-DFM efficiently estimates parameter updates caused by delayed conversions and new data, approximating retraining effects without duplicating samples or relying on auxiliary models. We further develop a scalable stochastic optimization algorithm to accelerate this computation. Our approach is applicable in both online and offline contexts. Extensive experiments validate that IF-DFM effectively mitigates the delayed feedback issue while incorporating new data to adjust to evolving user interests.
\section{Limitations and Future Work}
{While IF-DFM effectively addresses delayed feedback in CVR tasks, it relies on gradient information and assumes access to the full training set. Future work will enhance group influence estimation via higher-order Taylor expansions and explore causal inference to handle label reversals without auxiliary models. We also plan to extend our study by deploying the method in real-world settings with A/B testing. Furthermore, data sparsity remains a key challenge.}

\bibliography{dfm}

\end{document}